\pdfoutput=1
\documentclass[11pt]{article}

\usepackage{ACL2023}

\usepackage{times}
\usepackage{latexsym}
\usepackage{url}
\usepackage{graphicx}
\usepackage{multirow}
\usepackage{booktabs} % for professional tables
\usepackage{float}
\usepackage[T1]{fontenc}
\usepackage[utf8]{inputenc}

\usepackage{microtype}

\usepackage{inconsolata}
\usepackage{array, makecell} %
\usepackage{paralist}
\usepackage{enumitem}
\usepackage{arydshln}

\usepackage{titlesec}
\titlespacing{\paragraph}{%
  0pt}{%              left margin
  0.2\baselineskip}{% space before (vertical)
  1em}%

\usepackage{xspace}
\newcommand{\eg}{\hbox{\textit{e.g.,}}\xspace}

\usepackage{xfp}
\usepackage{anyfontsize}

\makeatletter
{\small % Capture font definitions of \small
\xdef\f@size@small{\f@size}
\xdef\f@baselineskip@small{\f@baselineskip}
\normalsize % Capture font definitions for \normalsize
\xdef\f@size@normalsize{\f@size}
\xdef\f@baselineskip@normalsize{\f@baselineskip}
}
\newcommand{\smalltonormalsize}{%
  \fontsize
    {\fpeval{(\f@size@small+\f@size@normalsize)/2}}
    {\fpeval{(\f@baselineskip@small+\f@baselineskip@normalsize)/2}}%
  \selectfont
}
\makeatother
  
\title{PLUE: Language Understanding Evaluation Benchmark \\ for Privacy Policies in English}
\author{
Jianfeng Chi$^{1,2}$ ~~ Wasi Uddin Ahmad$^{3}$\thanks{~~Work done while at UCLA.} ~~ Yuan Tian$^{3}$ ~~ Kai-Wei Chang$^{3}$ \\
$^{1}$Meta AI, $^{2}$University of Virginia, $^{3}$University of California, Los Angeles \\
\texttt{jianfengchi@meta.com,\{wasiahmad,yuant,kwchang\}@ucla.edu}
}

\begin{document}
\maketitle
\begin{abstract}
Privacy policies provide individuals with information about their rights and how their personal information is handled. Natural language understanding (NLU) technologies can support individuals and practitioners to understand better privacy practices described in lengthy and complex documents. However, existing efforts that use NLU technologies are limited by processing the language in a way exclusive to a single task focusing on certain privacy practices. To this end, we introduce the Privacy Policy Language Understanding Evaluation (PLUE) benchmark, a multi-task benchmark for evaluating the privacy policy language understanding across various tasks. We also collect a large corpus of privacy policies to enable privacy policy domain-specific language model pre-training. 
We evaluate several generic pre-trained language models and continue pre-training them on the collected corpus.
We demonstrate that domain-specific continual pre-training offers performance improvements across all tasks.
% and provide a series of analyses and ablation studies investigating important modeling considerations and challenges in PLUE.
The code and models are released at \url{https://github.com/JFChi/PLUE}.
% \footnote{ \url{https://github.com/JFChi/PLUE}}
\end{abstract}

\section{Introduction}

Privacy policies are documents that outline how a company or organization collects, uses, shares, and protects individuals' personal information.
% (e.g., name, address, phone number, and email address). 
% They allow users to make informed decisions about whether to provide their personal information to a company or organization.
Without a clear understanding of privacy policies, individuals may not know how their personal information is being used or who it is being shared with. The privacy violation might cause potential harm to them. 
% In addition, privacy policies help companies build trust with their users by demonstrating their commitment to protecting personal information and complying with laws and regulations related to data protection and privacy, such as the General Data Protection Regulation~\citep{voigt2017eu} in the European Union.
However, privacy policies are lengthy and complex, prohibiting users from reading and understanding them in detail \citep{federal2012protecting, gluck2016short, marotta2015does}. 
% This motivates the need for language technologies to improve understanding of privacy policies by processing them in a way that meets the needs of internet and mobile users.

Various natural language understanding (NLU) technologies have recently been developed to understand privacy policies \citep{wilson-etal-2016-creation, harkous2018polisis, ravichander-etal-2019-question, ahmad-etal-2020-policyqa, parvez2022retrieval, ahmad-etal-2021-intent, piextract_pets21}.
% , such as text classification~\citep{wilson-etal-2016-creation, harkous2018polisis}, question answering~\citep{ravichander-etal-2019-question, ahmad-etal-2020-policyqa, parvez2022retrieval}, semantic parsing~\citep{ahmad-etal-2021-intent}, and named-entity recognition~\citep{piextract_pets21}.
These tasks focus on understanding specific privacy practices at different syntax or semantics levels and require significant effort for data annotations (e.g., domain experts). It is hard to develop generic pre-trained language models (e.g., BERT~\citep{devlin-etal-2019-bert}) with task-specific fine-tuning using limited annotated data. Besides, the unique characteristics of privacy policies, such as reasoning over ambiguity and vagueness, modality, and document structure \citep{ravichander-etal-2021-breaking}, make it challenging to directly apply generic pre-trained language models to the privacy policy domain.

To address these problems and encourage research to develop NLU technologies in the privacy policy domain, we introduce the {\bf P}rivacy Policy {\bf L}anguage {\bf U}nderstanding {\bf E}valuation (PLUE) benchmark, 
% a multi-task benchmark for evaluating 
to evaluate the privacy policy language understanding across six tasks, including text classification, question answering, semantic parsing, and named-entity recognition. PLUE also includes a pre-training privacy policy corpus that we crawl from the websites to enable privacy policy domain-specific language model pre-training. We use this corpus to pre-train BERT~\citep{devlin-etal-2019-bert}, RoBERTa~\citep{liu2019roberta}, Electra~\citep{clark2020electra}, and SpanBERT~\citep{joshi-etal-2020-spanbert} and fine-tune them on the downstream tasks. 
We demonstrate that domain-specific continual pre-training offers performance improvements across all tasks.
% and provide a series of analyses and ablation studies highlighting important modeling considerations and challenges across tasks.
We will release the benchmark
%, including the pre-training codes and corpus and privacy-policy-related tasks and datasets, 
to assist natural language processing (NLP) researchers and practitioners in future exploration.
% of privacy-policy-related NLP.

\section{Policy Language Understanding Evaluation (PLUE) Benchmark}

\begin{table*}[!th]
\centering
% \resizebox{\textwidth}{!}{%
% \small
\smalltonormalsize
\begin{tabular}{l l l r r r r l}
\toprule
Dataset  & Task & Sub-domain & |Policy| & $|$Train$|$ & $|$Dev$|$ & $|$Test$|$ & Metric \\ \midrule
% \\ \midrule
OPP-115 & Classification & Websites & 115 & 2,771 & 395 & 625 & F1  \\
APP-350 & Classification & Mobile Apps & 350 & 10,150 & 2,817 & 2,540 & F1\\ \midrule
PrivacyQA & QA & Mobile Apps & 35 & 1,350 & -- & 400 & P / R / F1 \\
PolicyQA & QA & Mobile Apps & 115 & 17,056 & 3,809 & 4,152 & F1 / EM \\  \midrule
PolicyIE & \makecell[l]{Intent Classification\\Slot Filling} & \makecell[l]{Websites\\Mobile Apps} & 31 & 4,209 & -- & 1,041 & F1 / EM \\
\midrule
PI-Extract & NER & Websites & 30 & 3,034 & -- & 1,028 & F1 \\
\bottomrule
\end{tabular}%
% }
\vspace{-2mm}
\caption{Statistics of the PLUE datasets and tasks.
% \jianfeng{Shall we add a column indicating the text granularity, e.g., passage/text segments, sentences, words, of the annotations/input text?}
}
\vspace{-2mm}
\label{tab:data_stats}
\end{table*}

PLUE is centered on six English privacy policy language understanding tasks. 
The datasets and tasks are selected based on the following principles: 
(1) usefulness: the selected tasks can help practitioners in the domain quickly understand privacy practices without reading the whole privacy policy;
(2) task diversity: the selected tasks focus on different semantic levels, e.g., words (phrases), sentences, and paragraphs;
(3) task difficulty: the selected tasks should be adequately challenging for more room for improvement;
(4) training efficiency: all tasks can be trainable on a single moderate GPU (e.g., GeForce GTX 1080 Ti) for no more than ten hours;
(5) accessibility: all datasets are publicly available under licenses that allow usage and redistribution for research purposes.

\subsection{Datasets and Tasks}
PLUE includes six tasks in four categories. Table~\ref{tab:data_stats} presents an overview of the datasets and tasks within PLUE, and Table~\ref{tab:examples} in the Appendix gives an example for each task. % in PLUE.
% We describe each task as follows.

\paragraph{OPP-115} \citet{wilson-etal-2016-creation} presented 115 Online Privacy Policies (OPP-115). 
% The dataset comprises website privacy policies with annotations that specify privacy practices. Each text segment in privacy policies is annotated with one or more privacy practices from ten categories (e.g., \texttt{First Party Collection/Use}, \texttt{Data Retention}, \texttt{User Access, Edit and Deletion}).
The dataset comprises website privacy policies with text segments annotated with one or more privacy practices from ten categories (see Appendix \ref{appendix:opp115}).
% (e.g., \texttt{First Party Collection/Use}, \texttt{Data Retention}, \texttt{User Access, Edit and Deletion}). 
We train a multi-label classifier to predict the privacy practices given a sentence from a policy document. 
% The data in OPP-115 are class-imbalanced, and we use a macro (unweighted) average F1 for evaluation.  

\paragraph{APP-350} \citet{zimmeck2019maps} presented APP-350, a collection of mobile application privacy policies annotating what types of users' data mobile applications collect or share.
% data-type-related privacy practices.
% and could also be used for text classification. 
Like OPP-115, each text segment in a policy document is annotated with zero or more privacy practices (listed in Appendix \ref{appendix:app350}). 
% Still, APP-350 focuses on what types of data mobile applications collect or share from users. 
% For example, the privacy practices in APP-350 include \texttt{Demographic\_Age} and \texttt{Location\_GPS} (i.e., describes the collection/sharing of the user's age and GPS location data). 
In total, there are 30 data-type-related classes in APP-350, and we assign one more class, \texttt{No\_Mention}, to those text segments that do not pertain to such practices. 
% APP-350 data are skewed towards certain data types, and we again compute the unweighted average of the F1 score.

\paragraph{PrivacyQA} \citet{ravichander-etal-2019-question} proposed a question-answering dataset, PrivacyQA, comprised of 35 mobile application privacy policies. Given a question from a mobile application user and a sentence from a privacy policy, the task is to predict whether the sentence is relevant to the question. PrivacyQA includes unanswerable and subjective questions and formulates the QA task as a binary sentence classification task. 
% Following \citet{ravichander-etal-2019-question}, we evaluate with precision, recall, and F1 score.

\paragraph{PolicyQA} \citet{ahmad-etal-2020-policyqa} proposed a reading comprehension \citep{rajpurkar-etal-2016-squad} style dataset, PolicyQA. The dataset is derived from OPP-115 annotations that include a set of fine-grained attributes and evidence text spans that support the annotations. Considering the annotated spans as the answer spans, PolicyQA generates diverse questions relating to the corresponding privacy practices and attributes. The task is to predict the answer text span given the corresponding text segment and question. 
% We compute the F1 score and the exact match (EM) of predicted tokens versus the gold answer.

\paragraph{PolicyIE} \citet{ahmad-etal-2021-intent} proposed a semantic parsing dataset composed of two tasks: intent classification and slot filling. Given a sentence in a privacy policy, the task is to predict the sentence's intent (i.e., privacy practice) and identify the semantic concepts associated with the privacy practice. 
% For example, the sentence ``\textit{We do not sell your personal information to third parties}'' has the intent \textit{Data Sharing/Disclosure} and slots-value pairs (1) \textit{Data Sharer: First Party Entity}--\textit{We}, (2) \textit{Polarity: Negation}--\textit{not}, (3) \textit{Action}--\textit{sell}, (4) \textit{Data Shared: General Data}--\textit{personal information}, and (5) \textit{Data Receiver: Third Party Entity}--\textit{third parties}. 
Based on the role of the slots in privacy practices, PolicyIE groups them into type-I and type-II slots. In total, there are four intent labels and 14 type-I and four type-II slot labels. We individually train a text classifier and sequence taggers to perform intent classification and slot filling, respectively. 
% The evaluation metrics are the macro average of the F1 score for intent classification, token-level F1 and exact match for slot filling.

\paragraph{PI-Extract} \citet{piextract_pets21} presented PI-Extract, a named-entity recognition (NER) dataset. It aims to identify what types of user data are (not) collected or shared mentioned in the privacy policies. It contains 4 types of named entities: \texttt{COLLECT}, \texttt{NOT\_COLLECT}, \texttt{SHARE} and \texttt{NOT\_SHARE}. 
% As the name suggests, \texttt{COLLECT} and \texttt{NOT\_COLLECT} correspond to the types of data that are collected and not collected mentioned in the privacy policies.
% We compute the F1 score of the predicted entities versus the ground truth. 
Note that the named entities of different types may overlap. Thus, we report the results for collection-related and share-related entities, respectively.

\subsection{Pre-training Corpus Collection}
\label{subsec:pretrain_corpus}
The existing pre-trained language models (PLMs) mostly use data from BooksCorpus~\citep{zhu2015aligning} and English Wikipedia. Language models pre-trained on text from those sources might not perform well on the downstream privacy policy language understanding tasks, as privacy policies are composed of text written by domain experts (e.g., lawyers).
\citet{gururangan-etal-2020-dont} suggested that adapting to the domain's unlabeled data (domain-adaptive pre-training) improves the performance of domain-specific tasks.
Therefore, we collect a large privacy policy corpus for language model pre-training. In order to achieve broad coverage across privacy practices written in privacy policies \citep{william2020do, ahmad-etal-2021-intent}, we collect the privacy policies from two sources: mobile application privacy policies and website privacy policies. Appendix~\ref{appendix:pre-train-corpus-collection} provides more details about how we collect these two types of privacy policies.

% We use MAPS, the mobile application privacy policy corpus presented by \citet{zimmeck2019maps}. MAPS consists of the URLs of 441K mobile application privacy policies, which were collected from April to May 2018 from the Google Play store. We remove the duplicated URLs, crawl the privacy policy documents in HTML/PDF format, convert them to raw text format, and filter out the documents with noise (e.g., empty documents resulting from obsolete URLs). Finally, we ended up with 64K privacy policy documents. For website privacy policies, we use the Princeton-Leuven Longitudinal Corpus of Privacy Policies~\citep{amos2021privacy}.\footnote{The corpus is publicly available at \url{https://github.com/citp/privacy-policy-historical}.} The Princeton-Leuven Longitudinal Corpus of Privacy Policies contains 130K website privacy policies spanning over two decades. We use the documents with the latest date and convert them (from markdown format) into text format. Combining these two corpora, we obtain our pre-training corpus with 332M words.

\subsection{Models \& Training}

\begin{table*}[!th]
\centering
% \resizebox{0.8\linewidth}{!}{%
% \begin{minipage}
\smalltonormalsize
\begin{tabular}{l c c c c c c c}
\toprule
\multirow{2}{*}{Models} & \multirow{2}{*}{|Model|} & \textbf{OPP-115} & \textbf{APP-350} & \textbf{PrivacyQA} & \textbf{PolicyQA} & 
\textbf{PI-Extract} \\
\cdashline{3-7}
& & F1 & F1 & P / R / F1 & F1 / EM & F1 \\
\midrule
Human & - & - & - & 68.8 / 69.0 / 68.9 & - & - \\
\hdashline
BERT$_{\scriptsize \textsc{BASE}}$ & 110M & 75.3 & 59.6 & 44.6 / 35.9 / 36.3 & 55.1 / 27.7 & 63.7 / 54.6 \\
Electra$_{\scriptsize \textsc{BASE}}$ & 110M & 74.0 & 49.3 & 42.7 / 36.0 / 36.1 & 57.5 / 29.9 & 69.4 / 57.8 \\
SpanBERT$_{\scriptsize \textsc{BASE}}$ & 110M & 62.8 & 32.8 & 24.8 / 24.8 / 24.8 & 55.2 / 27.8 & 66.9 / 41.0 \\
RoBERTa$_{\scriptsize \textsc{BASE}}$ & 124M & 79.0 & 67.1 & 43.6 / 36.4 / 36.7 & 56.6 / 29.4 & 70.7 / 56.8 \\
\hdashline
PP-BERT$_{\scriptsize \textsc{BASE}}$ & 110M & 78.0 & 62.8 & 44.8 / 36.9 / 37.7 & 58.3 / 30.0 & 70.5 / 55.3  \\
PP-Electra$_{\scriptsize \textsc{BASE}}$ & 110M & 73.1 & 57.1 & 48.3 / 38.8 / 39.3 & 58.0 / 30.0 & 70.3 / 61.2  \\
PP-SpanBERT$_{\scriptsize \textsc{BASE}}$ & 110M & 78.1 & 61.9 & 43.4 / 36.4 / 36.8 & 55.8 / 27.5 & 65.5 / 50.8 \\
PP-RoBERTa$_{\scriptsize \textsc{BASE}}$ & 124M & {\bf 80.2} & 69.5 & {\bf 49.8} / {\bf 40.1} / {\bf 40.9} & 57.8 / 30.3 & {\bf 71.2} / 61.3 \\
\hdashline
LEGAL-BERT$_{\scriptsize \textsc{BASE}}$ & 110M & 76.0 & 57.4 & 45.6 / 37.6 / 38.2 & 55.1 / 27.7 & 69.1 / 51.1 \\
\hdashline
BERT$_{\scriptsize \textsc{LARGE}}$ & 340M & 79.3 & 71.2 & 43.8 / 35.4 / 36.1 & 56.6 / 28.7 & 68.1 / 54.8  \\
Electra$_{\scriptsize \textsc{LARGE}}$ & 340M & 78.7 & 41.5 & 46.6 / 42.1 / 40.5 & {\bf 60.7} / {\bf 33.2} & 70.1 / 59.5 \\
SpanBERT$_{\scriptsize \textsc{LARGE}}$ & 340M & 79.4 & 66.0 & 45.2 / 36.5 / 37.3 & 58.2 / 30.8 & 68.2 / 50.8 \\
RoBERTa$_{\scriptsize \textsc{LARGE}}$ & 355M & 79.9 & {\bf 72.4} & 47.6 / 41.4 / 40.6 & 59.8 / 32.5 & 70.9 / {\bf 62.8} \\
\bottomrule
\end{tabular}
% }
\vspace{-2mm}
\caption{
Performance comparison of pre-trained models on text classification, question answering, and named entity recognition tasks. We fine-tune all the models three times with different seeds and report average performances. Human performances are reported from the respective works.
}
\label{tab:benchmark-1}
% \end{minipage}
\bigskip
% \begin{minipage}
% \small
\smalltonormalsize
\begin{tabular}{l c c c c c c}
\toprule
\multirow{3}{*}{Models} & \multirow{3}{*}{|Model|}
& \textbf{Intent} & \multicolumn{4}{c}{\textbf{Slot Filling}} \\
& & \textbf{Classification} & \multicolumn{2}{c}{\textbf{Type-I Slots}} & \multicolumn{2}{c}{\textbf{Type-II Slots}}\\
\cdashline{3-7}
& & F1 & F1 & EM & F1 & EM \\
% \makecell{\textbf{PrivacyQA}\\F1} & \makecell{\textbf{Slot Filling}\\F1/EM} & \makecell{\textbf{PolicyIE}\\IC F1/ SF F1} & 
% \makecell{\textbf{PI-Extract}\\F1} \\
\midrule
Human & - & 96.5 & 84.3 & 56.6 & 62.3 & 55.6 \\
\hdashline
BERT$_{\scriptsize \textsc{BASE}}$ & 110M & 73.7	& 55.2 & 19.7 & 34.7 &	29.8 \\
Electra$_{\scriptsize \textsc{BASE}}$ & 110M & 73.7 & 56.4 & 22.8 & 36.5 & 30.7 \\
SpanBERT$_{\scriptsize \textsc{BASE}}$ & 110M & 71.9 & 44.0 & 10.8 & 29.7 & 17.5 \\
RoBERTa$_{\scriptsize \textsc{BASE}}$ & 110M & 74.5 & 56.8 & 22.0 & 39.2 & 32.0 \\
\hdashline
PP-BERT$_{\scriptsize \textsc{BASE}}$ & 110M & 76.9 & 56.7 & 22.8 & 38.7 & 32.5 \\
PP-Electra$_{\scriptsize \textsc{BASE}}$ & 110M & 77.1 & 58.2 & {\bf 24.1} & 37.8 & {\bf 32.9} \\
PP-SpanBERT$_{\scriptsize \textsc{BASE}}$ & 110M & 75.0 & 54.1 & 19.8 & 33.6 & 26.7 \\
PP-RoBERTa$_{\scriptsize \textsc{BASE}}$ & 110M & {\bf 78.1} & 58.0 & 22.4 & 40.1 & 32.4 \\
\hdashline
LEGAL-BERT$_{\scriptsize \textsc{BASE}}$ & 110M & 72.6 & 53.8 & 19.5 & 36.1 & 29.7 \\
\hdashline
BERT$_{\scriptsize \textsc{LARGE}}$ & 340M & 75.5 & 56.8 & 23.0 & 38.4 & 32.2 \\
Electra$_{\scriptsize \textsc{LARGE}}$ & 340M & 75.6 & 57.9 & 24.0 & 39.6 & 32.4 \\
SpanBERT$_{\scriptsize \textsc{LARGE}}$ & 340M & 73.8 & 45.5 & 9.5 & 38.8 & 29.8 \\
RoBERTa$_{\scriptsize \textsc{LARGE}}$ & 355M & 77.6 & {\bf 58.4} & 22.9 & {\bf 41.4} & 32.7 \\
\bottomrule
\end{tabular}
\vspace{-2mm}
\caption{
Performance comparison of pre-trained models on intent classification and slot filling tasks (PolicyIE). We fine-tune all the models three times with different seeds and report average performances. Human performances are reported from the respective works.
}
\vspace{-2mm}
\label{tab:benchmark-2}
% \end{minipage}
\end{table*}

% We benchmark a few pre-trained language models as baselines to facilitate future work.
\paragraph{Baselines}
We benchmark pre-trained language models (PLMs), BERT \cite{devlin-etal-2019-bert}, RoBERTa \cite{liu2019roberta}, SpanBERT \cite{joshi-etal-2020-spanbert}, Electra \cite{clark2020electra}, and LEGAL-BERT \cite{chalkidis-etal-2020-legal}. 
We present the details of the PLMs in Appendix~\ref{appendix:baseline}.

\paragraph{Domain-specific Continual Pre-training}
In order to adapt PLMs to the privacy policy domain, we continue to train BERT, Electra, SpanBERT, and RoBERTa on the pre-training corpus described in Section~\ref{subsec:pretrain_corpus}. We refer to them as PP-BERT, PP-RoBERTa, PP-SpanBERT, and PP-Electra, respectively.\footnote{We continually pre-train only the base models to mitigate the environmental impact of our experiments, but our code supports continual pre-training of large PLMs too.}
We present details in Appendix \ref{appendix:domain_training}.
% Since BERT, Electra, and SpanBERT share the same model architectures, we use almost the same hyperparameters (e.g., learning rate, train steps, batch size) for them following the original papers. 
% We scale down the train steps by the same factor, as the size of our pre-training corpus is roughly 1/10 the size of the pre-training corpus of BERT.
% We adhere to the guidelines outlined in \citet{liu2019roberta} to train RoBERTa with a larger batch size, higher learning rate, and fewer train steps.
% Table~\ref{tab:pretraining_hyperparams} in the Appendix presents the key hyperparameters we use to train the PLMs.

\paragraph{Task-specific Fine-tuning}
We fine-tune PLMs for each PLUE task. We \emph{only} tune the learning rate for each task, as we found in the preliminary experiments that model performances are highly sensitive to the learning rate.
% Table~\ref{tab:finetuning_hyperparams} in the Appendix lists the hyperparameters for all the downstream tasks.
We present more details in Appendix~\ref{appendix:task_finetuning}.
% gives more details of domain-specific continual pre-training and task-specific fine-tuning.

% Given the pre-trained language models, we fine-tune the models for each task using the Adam~\citep{kingma2015adam} optimizer with a batch size of 32. 
% We fine-tune the models on the QA tasks for 3 epochs and other tasks for 20 epochs and perform a grid search on the learning rate for each task.
% Tables~\ref{tab:finetuning_hyperparams} and~\ref{tab:learning_rate} in the Appendix list the hyperparameters for all the downstream tasks.

% In OPP-115 and APP-350, we compute the class weights (the class weights are inversely proportional to the occurrences of the classes) and apply them in fine-tuning, as we find out both datasets have the class-imbalance problem and using class weights brings gains to overall performance. We also report the human performances for PrivacyQA and PolicyIE from the original works.

% \subsection{Software Tools}  To facilitate using PLUE, we release our implementation, which is built with Pytorch~\cite{paszke2019pytorch} and the Huggingface transformers\footnote{\url{https://github.com/huggingface/transformers}} package. Our implementation includes the continual pre-training of our baselines and the evaluation of any PLMs supported by the Huggingface transformers package on the PLUE benchmark tasks. In addition to PLUE datasets, we release the pre-training corpus and all data pre-processing scripts, including the pre-training corpus crawling scripts, to assist future research in this area.

\section{Experiment Results}

% \subsection{Overall Results} 
Tables~\ref{tab:benchmark-1} and~\ref{tab:benchmark-2} present the results for all the experiment models for PLUE tasks. Rows 2-9 show the results of the base PLMs and their corresponding variants with privacy policy domain-specific continual pre-training. We observe that the language models (PP-BERT, PP-SpanBERT, PP-Electra, PP-RoBERTa) adapted to the privacy policy domain via continual pre-training outperform the general language models consistently in all the tasks. In particular, PP-RoBERTa performs the best for OPP-115, APP-350, PrivacyQA,\footnote{\citet{ravichander-etal-2019-question} reported 39.8\% F1 score for BERT model; however, we are able to achieve 36.3\%.} and PI-Extract, among all base models. PP-BERT and PP-RoBERTa perform the best for PolicyQA; PP-Electra and PP-RoBERTa achieve the best performance for PolicyIE.
In contrast, LEGAL-BERT (row 10) performs comparably or shows moderate improvements over BERT, indicating that pre-training on the general legal corpus does not necessarily help privacy policy language understanding.

It is interesting to see that continual pre-training of the language models using the privacy policy domain data benefits them differently. For example, in the text classification tasks (i.e., OPP-115 and APP-350), the performance difference between SpanBERT and PP-SpanBERT are most significant, while models using MLM (BERT and RoBERTa) already shows relatively high performance before continual pre-training and continual pre-training brings moderate gains to BERT and RoBERTa. 

We further investigate the improvement of large variants of PLMs over base variants of PLMs on PLUE tasks. As shown in the last four rows in Tables~\ref{tab:benchmark-1} and~\ref{tab:benchmark-2}, the large pre-trained language models mostly outperform their base counterparts (Exceptions include Electra for APP-350 and BERT for PrivacyQA). Noticeably, RoBERTa$_{\scriptsize \textsc{LARGE}}$ is the best-performing model in APP-350 and sub-tasks in PolicyIE and PI-Extract; Electra$_{\scriptsize \textsc{LARGE}}$ shows the highest F1 score and EM in PolicyQA.

Lastly, even though domain-specific pre-training and large PLMs help boost the performance for all tasks, the performance of some tasks and datasets (e.g., APP-350, PrivacyQA, slot filling in PolicyIE) remains low, which indicates much potential for further work on NLP for the privacy policy domain.

\section{Related Work}

% \paragraph{Language Evaluation Benchmarks}
% GLUE~\citep{wang2018glue}
% XTREME~\citep{pmlr-v119-hu20b}
% SuperGLUE~\citep{wang2019superglue}
% LexGLUE~\citep{chalkidis-etal-2022-lexglue}
% JGLUE~\citep{kurihara-etal-2022-jglue}
% BasqueGLUE~\citep{urbizu-etal-2022-basqueglue}
% NumGLUE~\citep{mishra-etal-2022-numglue}

\paragraph{Privacy Policy Benchmarks}
The Usable Privacy Policy Project \cite{sadeh2013usable} is the most significant effort to date, resulting in a large pool of works \cite{wilson-etal-2016-creation, wilson2016crowdsourcing, sathyendra2016automatic, mysore-sathyendra-etal-2017-identifying, bhatia2015towards, bhatia2016automated, hosseini2016lexical, zimmeck2019maps} to facilitate the automation of privacy policy analysis.
% Natural language processing (NLP) techniques including text alignment~\citep{liu-etal-2014-step, ramanath-etal-2014-unsupervised}, text classification~\citep{wilson-etal-2016-creation, harkous2018polisis, zimmeck2019maps}, question answering (QA)~\citep{shvartzshanider-etal-2018-recipe, harkous2018polisis, ravichander-etal-2019-question, ahmad-etal-2020-policyqa}, named entity recognition \cite{piextract_pets21}, and semantic parsing \cite{ahmad-etal-2021-intent} have been explored. 
A wide range of NLP techniques have been explored accordingly \citep{liu-etal-2014-step, ramanath-etal-2014-unsupervised, wilson-etal-2016-creation, harkous2018polisis, zimmeck2019maps, shvartzshanider-etal-2018-recipe, harkous2018polisis, ravichander-etal-2019-question, ahmad-etal-2020-policyqa, piextract_pets21, ahmad-etal-2021-intent}. 
% In this work, we aggregate some of these efforts under one umbrella and benchmark pre-trained language models.

\paragraph{Pre-trained Language Models} 
In the last few years, NLP research has witnessed a radical change with the advent of PLMs like ELMo \citep{peters-etal-2018-deep} and BERT \citep{devlin-etal-2019-bert}. 
PLMs achieved state-of-the-art results in many language understanding benchmarks. 
% In particular, PLMs found to be very effective in low-resource settings. 
Consequently, PLMs have been developed for a wide range of domains, \eg scientific \cite{beltagy-etal-2019-scibert}, medical \cite{lee2020biobert, rasmy2021med, alsentzer-etal-2019-publicly}, legal \cite{chalkidis-etal-2020-legal}, and cybersecurity \cite{ranade2021cybert, bayer2022cysecbert}. This work investigates the adaptation of PLMs to facilitate NLP research in the privacy policy domain.

% \paragraph{Pre-trained Language Models} 
% In the last few years, NLP research has witnessed a radical change with the advent of pre-trained language models (PLMs) like ELMo \citep{peters-etal-2018-deep} and BERT \citep{devlin-etal-2019-bert}. 
% PLMs achieved state-of-the-art results in many language understanding benchmarks. 
% % In particular, PLMs found to be very effective in low-resource settings. 
% Consequently, PLMs have been developed for a wide range of domains, giving birth to SciBERT \cite{beltagy-etal-2019-scibert} for scientific NLP, BioBERT \cite{lee2020biobert}, Med-BERT \cite{rasmy2021med}, and ClinicalBERT \cite{alsentzer-etal-2019-publicly} for medical NLP, LEGAL-BERT \cite{chalkidis-etal-2020-legal} for legal NLP, CyBERT \cite{ranade2021cybert} and CySecBERT \cite{bayer2022cysecbert} for Cybersecurity domain. This work particularly investigates the adaptation of PLMs to facilitate NLP research in the privacy policy domain.

\section{Conclusion}
% Reliable aggregation of datasets benefits NLP research for low-resource domains. Benchmarking foundation models on these datasets could facilitate future research.
Reliable aggregation of datasets and benchmarking foundation models on them facilitate future research.
This work presents PLUE, a benchmark for training and evaluating new security and privacy policy models.
PLUE will help researchers benchmark policy language understanding under a unified setup and facilitate reliable comparison.

% \jianfeng{Add more future work here, e.g., (1) unified multi-task learning using generative model (e.g., BART and T5), (2) compute human performance for other tasks, (3) robustness (e.g., how  robust the PLMs are under the PLUE task). (4) meta-learning for low-resource domain adaptation; Low-resource and few-shot learning}

\section*{Limitations}
The pre-training privacy policy corpus and the downstream task datasets are unlikely to contain toxic or biased content. Therefore, they should not magnify toxicity or bias in the pre-trained and fine-tuned models, although the models may exhibit such behavior due to their original pre-training. 
The pre-training and benchmark datasets are formed based on privacy policies crawled in the past; as a result, they could be outdated by now.
This work focuses on the English language only, and the findings may not apply to other languages.

\section*{Ethics Statement}

\paragraph{License} The OPP-115 and APP-350 datasets are made available for research, teaching, and scholarship purposes only, with further parameters in the spirit of a Creative Commons Attribution-NonCommercial License (CC BY-NC). The PolicyQA and PI-Extract datasets are derived from OPP-115 datasets. The PrivacyQA and PolicyIE dataset are released under an MIT license. The pre-training corpus, MAPS Policies Dataset, is released under CC BY-NC. We strictly adhere to these licenses and will release the PLUE benchmark resources under CC BY-NC-SA 4.0.

\paragraph{Carbon Footprint} We avoided using large models for continual training on the privacy policy domain, reducing their environmental impacts. The PP-BERT, PP-SpanBERT, and PP-Electra models were trained for 100k steps on Tesla V100 GPUs that took 1-2 days. Therefore, the training would emit only 9kg of carbon into the environment.\footnote{Calculated using \url{https://mlco2.github.io/impact}, based on a total of 100 hours of training on Tesla V100 and Amazon Web Services as the provider.} All fine-tuning experiments were very lightweight due to the small size of the datasets, resulting in approximately 12kg of carbon emission.

\section*{Acknowledgements}
We thank the anonymous reviewers for their insightful comments. This work was supported in part by National Science Foundation Grant OAC 2002985, OAC 1920462 and CNS 1943100, Google Research Award, and Meta Research Award. Any opinions, findings, conclusions, or recommendations expressed herein are those of the authors and do not necessarily reflect those of the US Government or NSF.

\bibliography{anthology,custom}
\bibliographystyle{acl_natbib}

\clearpage
\appendix
% \section*{Appendix}

\twocolumn[{%
 \centering
 \Large\bf Supplementary Material: Appendices \\ [20pt]
}]

\section{Dataset Details}

\subsection{OPP-115 Privacy Practices}
\label{appendix:opp115}

\begin{compactenum}
    \item First Party Collection/Use
    \item Third Party Sharing/Collection
    \item User Choice/Control 
    \item User Access, Edit, and Deletion
    \item Data Retention
    \item Data Security
    \item Policy Change
    \item Do Not Track 
    \item International and Specific Audiences
    \item Other
\end{compactenum}

\subsection{APP-350 Privacy Practices}
\label{appendix:app350}

% \begin{compactenum}
%     \item Contact 1st Party
%     \item Contact 3rd Party
%     \item Contact Email Address 1st Party
%     \item Contact Email Address 3rd Party
%     \item Contact Phone Number 1st Party
%     \item Contact Phone Number 3rd Party
%     \item Identifier 1st Party
%     \item Identifier 3rd Party
%     \item Identifier Cookie 1st Party
%     \item Identifier Cookie 3rd Party
%     \item Identifier Device ID 1st Party
%     \item Identifier Device ID 3rd Party
%     \item Identifier IMEI 1st Party
%     \item Identifier IMEI 3rd Party
%     \item Identifier IMSI 1st Party
%     \item Identifier IMSI 3rd Party
%     \item Identifier MAC 1st Party
%     \item Identifier MAC 3rd Party
%     \item Identifier Mobile Carrier 1st Party
%     \item Identifier Mobile Carrier 3rd Party
%     \item Identifier SIM Serial 1st Party
%     \item Identifier SIM Serial 3rd Party
%     \item Identifier SSID BSSID 1st Party
%     \item Identifier SSID BSSID 3rd Party
%     \item Location 1st Party
%     \item Location 3rd Party
%     \item Location Cell Tower 1st Party
%     \item Location Cell Tower 3rd Party
%     \item Location GPS 1st Party
%     \item Location GPS 3rd Party
%     \item Location WiFi 1st Party
%     \item Location WiFi 3rd Party
%     \item Single Sign On
%     \item Single Sign On: Facebook
% \end{compactenum}

\begin{compactenum}
    \item Contact
    \item Contact\_Address\_Book
    \item Contact\_City
    \item Contact\_E\_Mail\_Address
    \item Contact\_Password
    \item Contact\_Phone\_Number
    \item Contact\_Postal\_Address
    \item Contact\_ZIP
    \item Demographic
    \item Demographic\_Age
    \item Demographic\_Gender
    \item Facebook\_SSO
    \item Identifier
    \item Identifier\_Ad\_ID
    \item Identifier\_Cookie\_or\_similar\_Tech
    \item Identifier\_Device\_ID
    \item Identifier\_IMEI
    \item Identifier\_IMSI
    \item Identifier\_IP\_Address
    \item Identifier\_MAC
    \item Identifier\_Mobile\_Carrier
    \item Identifier\_SIM\_Serial
    \item Identifier\_SSID\_BSSID
    \item Location
    \item Location\_Bluetooth
    \item Location\_Cell\_Tower
    \item Location\_GPS
    \item Location\_IP\_Address
    \item Location\_WiFi
    \item SSO
\end{compactenum}

\begin{table*}[!ht]
\centering 
\resizebox{\linewidth}{!}{%
% \small
\begin{tabular}{p{0.10\textwidth}p{0.9\textwidth}}

\toprule
% \parbox[t]{1mm}{\multirow{2}{*}{\rotatebox[origin=c]{90}{{\textbf{OPP-115}}}}} 
\textbf{OPP-115}
&
\textbf{Text:} \textit{Secure Online Ordering \,\,   For your security, we only store your credit card information if you choose to set up an authorized account with one of our Sites. In that case, it is stored on a secure computer in an encrypted format. If you do not set up an account, you will have to enter your credit card information each time you order. We understand that this may be a little inconvenient for you, but some customers appreciate the added security.} \\ \\& \textbf{Classes:} \texttt{Data Security}, \texttt{User Choice/Control}, \texttt{First Party Collection/Use}\\

\midrule
% \parbox[t]{1mm}{\multirow{2}{*}{\rotatebox[origin=c]{90}{{\textbf{APP-350}}}}} 
\textbf{APP-350}
&
\textbf{Text:} \textit{Our Use of Web Beacons and Analytics Services Microsoft web pages may contain electronic images known as web beacons (also called single-pixel gifs) that we use to help deliver cookies on our websites, count users who have visited those websites and deliver co-branded products. We also include web beacons in our promotional email messages or newsletters to determine whether you open and act on them.} \\ \\ & \textbf{Classes:} \texttt{Contact\_E\_Mail\_Address}, \texttt{Identifier\_Cookie\_or\_similar\_Tech}\\

% \midrule
% \parbox[t]{1mm}{\multirow{2}{*}{\rotatebox[origin=c]{90}{{\textbf{PrivacyQA}}}}} & \textbf{Premise:} \textit{My body cast a shadow over the grass.}\quad
% \textbf{Question:} \textit{What’s the CAUSE for this?}\\
% &\textbf{Alternative 1:} \textit{The sun was rising.}\quad
% \textbf{Alternative 2:} \textit{The grass was cut.}\\
% &\textbf{Correct Alternative:} \texttt{1}\\

\midrule
% \parbox[t]{1mm}{\multirow{2}{*}{\rotatebox[origin=c]{90}{{\textbf{PrivacyQA}}}}} 
\textbf{PrivacyQA}
&
\textbf{Sentence:} \textit{We may collect and use information about your location (such as your country) or infer your approximate location based on your IP address in order to provide you with tailored educational experiences for your region, but we don't collect the precise geolocation of you or your device.}\\ \\
& \textbf{Question:} \textit{Does the app track my location?} \textbf{Answer:} \texttt{Relevant}\\

\midrule
% \parbox[t]{1mm}{\multirow{2}{*}{\rotatebox[origin=c]{90}{{\textbf{PolicyQA}}}}} 
\textbf{PolicyQA}
& 
\textbf{Text:} \textit{Illini Media never shares personally identifiable information provided to us online in ways unrelated to the ones described above without allowing you to opt out or otherwise prohibit such unrelated uses. Google or any ad server may use information (not including your name, address, email address, or telephone number) about your visits to this and other websites in order to provide advertisements about goods and services of interest to you.}\\ \\
&\textbf{Question:} Do you share my data with others? If yes, what is the type of data? \\
&\textbf{Answer:} \textit{information (not including your name, address, email address or telephone number)} \\

\midrule
% \parbox[t]{5mm}{\multirow{2}{*}{\rotatebox[origin=c]{90}{{\textbf{PolicyIE}}}}} 
\textbf{PolicyIE}
&
\textbf{Sentence:} \textit{We may also use or display your username and icon or profile photo for marketing purposes or press releases.}\\ \\
& \textbf{Intent:} \texttt{Data Collection/Usage} \quad
\textbf{Slots:} (1) \textit{Data Collector: First Party Entity--We}, (2) \textit{Action--use}, (3) \textit{Data Provider: User--your}, (4) \textit{Data Collected: User Online Activities/Profiles--username}, (5) \textit{Data Collected: User Online Activities/Profiles--icon or profile photo}, (6) \textit{Purpose: Advertising/Marketing--marketing purpose or press releases}. \\

% \midrule
% \parbox[t]{1mm}{\multirow{2}{*}{\rotatebox[origin=c]{90}{{\textbf{WiC}}}}} &
% \textbf{Context 1:} \textit{Room and \underline{board}.} \quad
% \textbf{Context 2:} \textit{He nailed \underline{boards} across the windows.} \\
% & \textbf{Sense match:} \texttt{False}\\

\midrule
% \parbox[t]{1mm}{\multirow{1}{*}{\rotatebox[origin=c]{90}{{\textbf{PI-Extract}}}}} 
\textbf{PI-Extract}
& 
\textbf{Text:} \textit{ We may share aggregate demographic and usage information with our prospective and actual business partners, advertisers, and other third parties for any business purpose.} \\\\
& \textbf{Entities:} \texttt{SHARE} -- \textit{aggregate demographic and usage information}\\
\bottomrule
\end{tabular}
}
\vspace{-2mm}
\caption{Examples from the tasks in PLUE. 
% \textbf{Bold} text represents part of the example format for each task. Text in \textit{italics} is part of the model input. \underline{\textit{Underlined}} text is specially marked in the input. Text in a \texttt{monospaced font} represents the expected model output.
}
\vspace{-2mm}
\label{tab:examples}
\end{table*}

\section{More Details of Pre-training Corpora}
\label{appendix:pre-train-corpus-collection}

We use MAPS, the mobile application privacy policy corpus presented by \citet{zimmeck2019maps}. MAPS consists of the URLs of 441K mobile application privacy policies, which were collected from April to May 2018 from the Google Play store. We remove the duplicated URLs, crawl the privacy policy documents in HTML/PDF format, convert them to raw text format, and filter out the documents with noise (e.g., empty documents resulting from obsolete URLs). Finally, we ended up with 64K privacy policy documents. For website privacy policies, we use the Princeton-Leuven Longitudinal Corpus of Privacy Policies~\citep{amos2021privacy}.\footnote{The corpus is publicly available at \url{https://github.com/citp/privacy-policy-historical}.} The Princeton-Leuven Longitudinal Corpus of Privacy Policies contains 130K website privacy policies spanning over two decades. We use the documents with the latest date and convert them (from markdown format) into text format. Combining these two corpora, we obtain our pre-training corpus with 332M words.

\section{Baseline Models}
\label{appendix:baseline}

We benchmark a few pre-trained language models as baselines to facilitate future work.

% \smallskip
% \noindent
% \textbf{$\bullet$ BERT \hspace{0.1em}} 
\paragraph{BERT}
\citet{devlin-etal-2019-bert} proposed Transformer \cite{vaswani2017attention} based language model pre-trained on BooksCorpus and English Wikipedia data using masked language modeling (MLM) and next sentence prediction. 

% \smallskip
% \noindent
% \textbf{$\bullet$ Electra \hspace{0.1em}} 
\paragraph{Electra}
\citet{clark2020electra} pre-trains a generator and a discriminator on the same corpus as BERT, where the generator takes a masked text as input and is trained using the MLM objective. The discriminator takes the predictions from the generator and detects which tokens are replaced by the generator. After pre-training, the generator is discarded, and the discriminator is used as the language model for the downstream tasks. 

% \smallskip
% \noindent
% \textbf{$\bullet$ SpanBERT \hspace{0.1em}} 
\paragraph{SpanBERT}
\citet{joshi-etal-2020-spanbert} shares the same architecture and pre-training corpus as BERT but differs in the pre-training objectives. It extends BERT by masking contiguous spans instead of single tokens and training the span boundary representations to predict the masked spans.

\paragraph{RoBERTa}
\citet{liu2019roberta} presented a replication study of BERT pretraining where they showed that BERT was significantly undertrained and proposed RoBERTa that tunes key hyperparameters and uses more training data to achieve remarkable performance improvements. Note that while BERT, Electra, and SpanBERT use the same vocabulary, RoBERTa uses a different vocabulary resulting in 15M more parameters in the model.

% \smallskip
% \noindent
% \textbf{$\bullet$ LEGAL-BERT} 
\paragraph{LEGAL-BERT}
\citet{chalkidis-etal-2020-legal} pre-trained BERT using 12 GB of the English text (over 351K documents) from several legal fields (e.g., contracts, legislation, court cases) scraped from publicly available resources. Since privacy policies serve as official documents to protect the company and consumers' privacy rights and might contain contents in response to privacy law (e.g., GDPR), we study LEGAL-BERT's effectiveness on the PLUE tasks.

\section{More Implementation Details}
\label{appendix:implementations}

% \wasi{We use class weights to tackle the label imbalance issue in OPP115 and APP350 text classification tasks.}
% \jianfeng{I mentioned it in the main text in the section ``Task-specific Fine-tuning''.}

\subsection{Domain-specific Continual Pre-training} 
\label{appendix:domain_training}
Since BERT, Electra, and SpanBERT share the same model architectures, we use almost the same hyperparameters (e.g., learning rate, train steps, batch size) for them following the original papers. 
We scale down the train steps by the same factor, as the size of our pre-training corpus is roughly 1/10 the size of the pre-training corpus of BERT.
We adhere to the guidelines outlined in \citet{liu2019roberta} to train RoBERTa with larger batch size, higher learning rate, and fewer train steps.
Table~\ref{tab:pretraining_hyperparams} 
% in the Appendix 
presents the training hyperparameters for PLMs.
% we use to train the PLMs.

\subsection{Task-specific Fine-tuning} 
\label{appendix:task_finetuning}
We fine-tune the models for each task using the Adam~\citep{kingma2015adam} optimizer with a batch size of 32. 
We fine-tune the models on the QA tasks for 3 epochs and other tasks for 20 epochs and perform a grid search on the learning rate for each task with validation examples. We chose the learning rate for tasks without validation examples based on our findings from the tasks with validation examples.
Table~\ref{tab:finetuning_hyperparams} 
% and~\ref{tab:learning_rate} 
lists the hyperparameters for all the downstream tasks.

In OPP-115 and APP-350, we compute the class weights (the class weights are inversely proportional to the occurrences of the classes) and apply them in fine-tuning, as we find out both datasets have the class-imbalance problem and using class weights brings gains to overall performance. We also report the human performances for PrivacyQA and PolicyIE from the original works.

\subsection{Software Tools}  To facilitate using PLUE, we release our implementation, which is built with Pytorch~\cite{paszke2019pytorch} and the Huggingface transformers\footnote{\url{https://github.com/huggingface/transformers}} package. Our implementation includes the continual pre-training of our baselines and the evaluation of any PLMs supported by the Huggingface transformers package on the PLUE benchmark tasks. In addition to PLUE datasets, we release the pre-training corpus and all data pre-processing scripts, including the pre-training corpus crawling scripts, to assist future research in this area.

\begin{table*}[!th]
    % \small
    \centering
    % \resizebox{\linewidth}{!}{
    \begin{tabular}{l c c c c }
    \toprule
    & \textbf{PP-BERT} & \textbf{PP-SpanBERT} & \textbf{PP-Electra} & \textbf{PP-RoBERTa}\\ \midrule
    Learning Rate & 1e-4 & 1e-4 & 1e-4  & 6e-4\\
    Train Steps & 100,000 & 100,000 & 100,000 & 12,500 \\
    batch Size & 256 & 256 & 256 & 2048 \\
    Learning Rate Schedule & linear & polynomial\_decay & linear & linear \\
    \# warm-up steps & 1000 & 1000 & 1000 & 600 \\
    Optimizer & AdamW & AdamW & AdamW & AdamW \\
    \bottomrule
    \end{tabular}
    % }
    \caption{Hyperparameters for pre-training language models.}
    \label{tab:pretraining_hyperparams}
    \bigskip
    \begin{tabular} {l  c  c  c  c  } % {l | c | c | c | c | c | c | c | c }
    \toprule
    & \makecell{\textbf{Text}\\\textbf{Classification}}  & \makecell{\textbf{Question}\\\textbf{Answering}} & \makecell{\textbf{Semantic}\\\textbf{Parsing}} &\makecell{\textbf{NER}} \\
    \midrule
    Dropout & 0.1 & 0.1 & 0.1 & 0.1 \\
    Weight decay & 0.0 & 0.0 & 0.0 & 0.0 \\
    Optimizer & AdamW & AdamW & AdamW & AdamW \\
    Batch Size & 32 & 32 & 32 & 32 \\
    Learning rate & \multicolumn{4}{c}{[3e-4, 1e-4, 5e-5, 3e-5, 1e-5, 5e-6, 3e-6]} \\
    Learning Rate Schedule & Linear & Linear & Linear & Linear \\
    Warm-up Ratio & 0.05 & 0.0 & 0.05 & 0.05 \\
    \# epoch & 20 & 3 & 20 & 20 \\
    \bottomrule
    \end{tabular}
    % }
    \caption{Hyperparameters for fine-tuning pre-trained language models on different PLUE tasks.}
    \label{tab:finetuning_hyperparams}
\end{table*}

% \section{Performance Analyses}

% \input{tables/appendix/breakdown.tex}
% \include{tables/appendix/full_result.tex}

\end{document}